\def\year{2021}\relax
\documentclass[letterpaper]{article} 
\usepackage{aaai21}  
\usepackage{times}  
\usepackage{helvet} 
\usepackage{courier}  
\usepackage[hyphens]{url}  
\usepackage{graphicx} 
\usepackage{natbib}  
\usepackage{caption} 
\usepackage{algorithm}
\usepackage{algorithmic}
\usepackage{multirow}
\usepackage{bm}
\newcommand{\tabincell}[2]{\begin{tabular}{@{}#1@{}}#2\end{tabular}}
\title{Leveraging Table Content for Zero-shot Text-to-SQL with Meta-Learning}
\author{
	
	Yongrui Chen,\textsuperscript{\rm 1}
	Xinnan Guo,\textsuperscript{\rm 1}
	Chaojie Wang,\textsuperscript{\rm 2}
	Jian Qiu,\textsuperscript{\rm 2}
	Guilin Qi,\textsuperscript{\rm 1}
	Meng Wang,\textsuperscript{\rm 1}
	Huiying Li\textsuperscript{\rm 1}
	\\
}
\affiliations{
	
	\textsuperscript{\rm 1}School of Computer Science and Engineering, Southeast University, Nanjing, China\\
	\textsuperscript{\rm 2}Alibaba Group\\
	
	 220171664@seu.edu.cn, guoxinnan0727@163.com, \{chaojie.wcj, qiujian.qj\}@alibaba-inc.com,\\ \{gqi, meng.wang, huiyingli\}@seu.edu.cn
	
}

\begin{document}
	
	\maketitle
	
	\begin{abstract}
	Single-table text-to-SQL aims to transform a natural language question into a SQL query according to one single table.  Recent work has made promising progress on this task by pre-trained language models and a multi-submodule framework. However, zero-shot table, that is, the invisible table in the training set, is currently the most critical bottleneck restricting the application of existing approaches to real-world scenarios. Although some work has utilized auxiliary tasks to help handle zero-shot tables, expensive extra manual annotation limits their practicality. In this paper, we propose a new approach for the zero-shot text-to-SQL task which does not rely on any additional manual annotations. Our approach consists of two parts. First, we propose a new model that leverages the abundant information of table content to help establish the mapping between questions and zero-shot tables. Further, we propose a simple but efficient meta-learning strategy to train our model. The strategy utilizes the two-step gradient update to force the model to learn a generalization ability towards zero-shot tables. We conduct extensive experiments on a public open-domain text-to-SQL dataset WikiSQL and a domain-specific dataset ESQL. Compared to existing approaches using the same pre-trained model, our approach achieves significant improvements on both datasets. Compared to the larger pre-trained model and the tabular-specific pre-trained model, our approach is still competitive. More importantly, on the zero-shot subsets of both the datasets, our approach further increases the improvements.
	\end{abstract}
	
	\section{Introduction}
	Since the release of WikiSQL~\cite{DBLP:journals/corr/abs-1709-00103}, a large-scale text-to-SQL benchmark, single-table text-to-SQL task has become an active research area in recent years. The goal of the task is to transform natural language questions into Structured Query Language (SQL) to query a single table. Although the search space is limited to one table, the task still has a considerable number of application scenarios (e.g., query regional electricity prices or flight schedules). More importantly, it is the basis for more complex text-to-SQL tasks on multi-tables~\cite{DBLP:conf/emnlp/YuZYYWLMLYRZR18}. Therefore, the research on this area is of great significance.
	
	Relying on large-scale pre-trained language models~\cite{DBLP:conf/naacl/DevlinCLT19} and a multi-submodule framework, existing approaches~\cite{DBLP:journals/corr/abs-1908-08113,DBLP:journals/corr/abs-1902-01069,DBLP:journals/corr/abs-2008-04759} have made considerable progress on the single-table text-to-SQL task. However, few of them pay attention to the challenge of zero-shot tables whose \textit{schema} are not visible in the training set. Typically, in comparison with the visible tables, zero-shot tables are more challenging because they are not directly involved in training. Their schema cannot be perceived by the model, so that they may be noisy in the test. In fact, with the rapid expansion of business, zero-shot tables are becoming more and more common in realistic scenarios. Therefore, in order to make text-to-SQL land from laboratory to application, it is necessary to make the model learn to handle zero-shot tables.
	
	\cite{DBLP:conf/aaai/ChangLT0HZ20} explicitly deals with zero-shot tables for the first time. The core idea of their approach is to design an auxiliary task to model the mapping from the question to the headers (similar to entity linking). However, this approach requires training data to provide the gold mappings that are annotated manually. Undoubtedly, it is a strong limitation in realistic scenarios.
	
	\begin{figure}
		\includegraphics[width=0.48\textwidth]{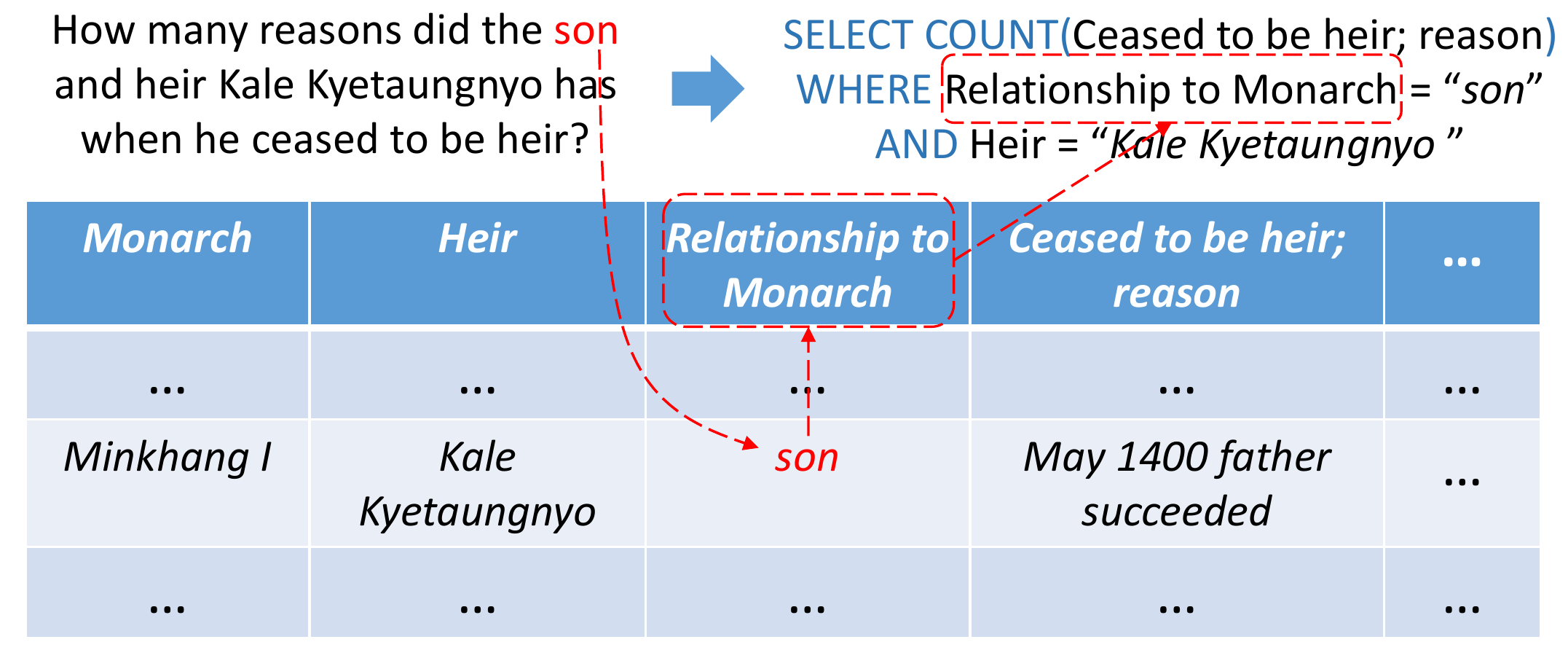}
		\caption{An example of table content to help predict headers. Red indicates the matching.} \label{fig_example}
	\end{figure}
	
	In this paper, we propose a new approach called \textit{Meta-Content text-to-SQL} (MC-SQL) to handle zero-shot tables. The motivation comes from the following two intuitions: 1) The first one is that table content can provide abundant information for predicting headers. Figure \ref{fig_example} shows an example. The cell \textit{son} in the table is relevant to the question word ``son", thus reveals the potential header \textit{Relationship to Monarch}. 2) The second one is that meta-learning can help the model learn the generalization ability between different tables from the training data. It is because meta-learning has the capability that only needs a few gradient steps to quickly adapt to new tasks. Specifically, our approach consists of two parts. On the one hand, a table content-enhanced model is employed to encode questions, headers, and table cells at the same time, in order to combine their semantic relevance for the prediction on zero-shot tables. On the other hand, a zero-shot meta-learning algorithm is utilized to train our content-enhanced model instead of the traditional mini-batch strategy. In each training step, the algorithm generalizes the model by two sets of samples that rely on two disjoint table sets, respectively. Finally, to comprehensively evaluate our approach, we conduct experiments on public open-domain benchmark WikiSQL and domain-specific benchmark ESQL. Our approach achieves a significant improvement over the baselines that utilizes the same pre-trained model as ours, and also achieves competitive results over the baselines that utilize the larger or tabular-specific pre-trained model.

	\section{Preliminaries}
	The single-table text-to-SQL task can be formally defined as
	\begin{equation}
	y = \mathcal{M}(q, \mathcal{T})
	\label{eq_task}
	\end{equation}
	where $q$ denotes a natural language question and $y$ denotes the corresponding SQL query. $\mathcal{T}=\{h^1,h^2,...,h^l\}$ denotes the table which $q$ relies on, where $h^i$ denotes the $i$-th header in $\mathcal{T}$. The goal of the task is to learn a mapping $\mathcal{M}$ from questions to SQL queries. In addition, this task supposes that no complex SQL syntax (e.g., GROUP BY and nested query) exists and there is only one column in the SELECT clause. Specifically, each $y$ follows a unified skeleton, which is shown in Figure \ref{fig_skeleton}. The tokens prefixed with ``\$" indicate the slots to be filled and ``*" indicates zero or more AND clauses. According to the skeleton, existing approaches~\cite{DBLP:journals/corr/abs-1908-08113,DBLP:journals/corr/abs-1902-01069,DBLP:journals/corr/abs-2008-04759} break the total task into the following six subtasks:
	
	\begin{figure}
		\includegraphics[width=0.48\textwidth]{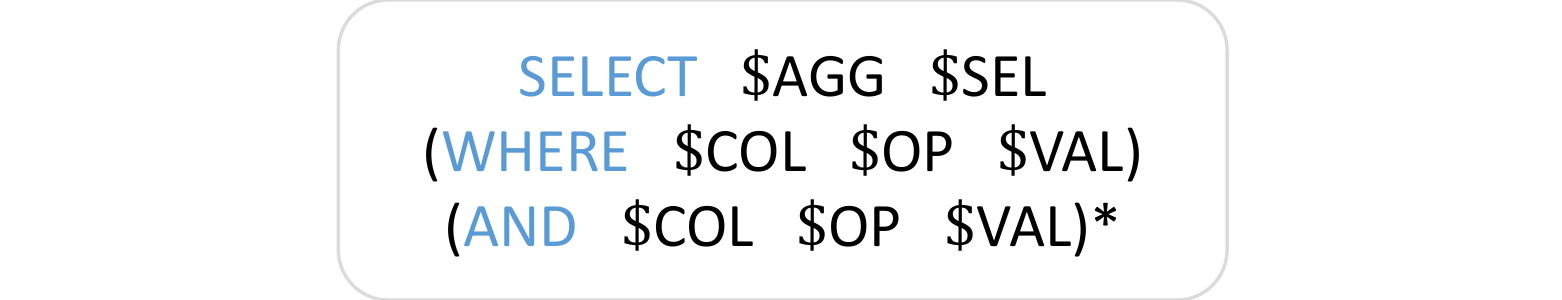}
		\caption{Skeleton of SQL in single-table text-to-SQL.} \label{fig_skeleton}
	\end{figure}
	
	\begin{itemize}
		\item \textbf{Select-Column(SC)} finds the column \$SEL in the SELECT clause from $\mathcal{T}$.
		\item \textbf{Select-Aggregation(SA)} finds the aggregation function \$AGG ($\in $ \{NONE, MAX, MIN, COUNT, SUM, AVG\}) of the column in the SELECT clause.
		\item \textbf{Where-Number(WN)} finds the number of where conditions, denoted by $\mathcal{N}$.
		\item \textbf{Where-Column(WC)} finds the column (header) \$COL of each WHERE condition from $\mathcal{T}$.
		\item \textbf{Where-Operator(WO)} finds the operator \$OP ($\in \{=,>,<\}$) of each \$COL in the WHERE clause.
		\item \textbf{Where-Value(WV)} finds the value \$VAL for each condition from the question, specifically, locating the starting position of the value in $q$.
	\end{itemize}

	There are dependencies between some tasks. For example, the prediction of \$OP requires \$COL, and the prediction of \$VAL requires both \$COL and \$OP.
	
	\begin{figure}
		\includegraphics[width=0.48\textwidth]{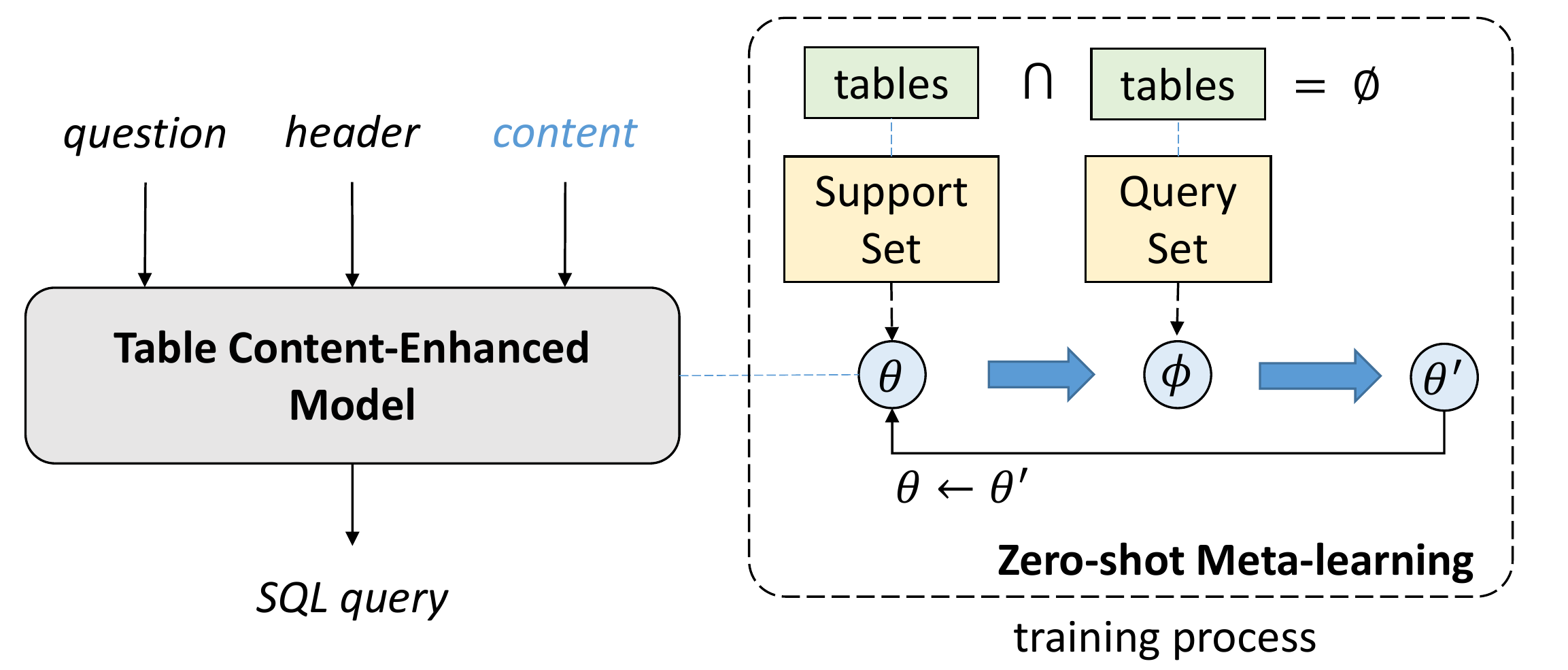}
		\caption{Overall framework of our approach.} \label{fig_overview}
	\end{figure}
	
	\begin{figure*}
		\includegraphics[width=\textwidth]{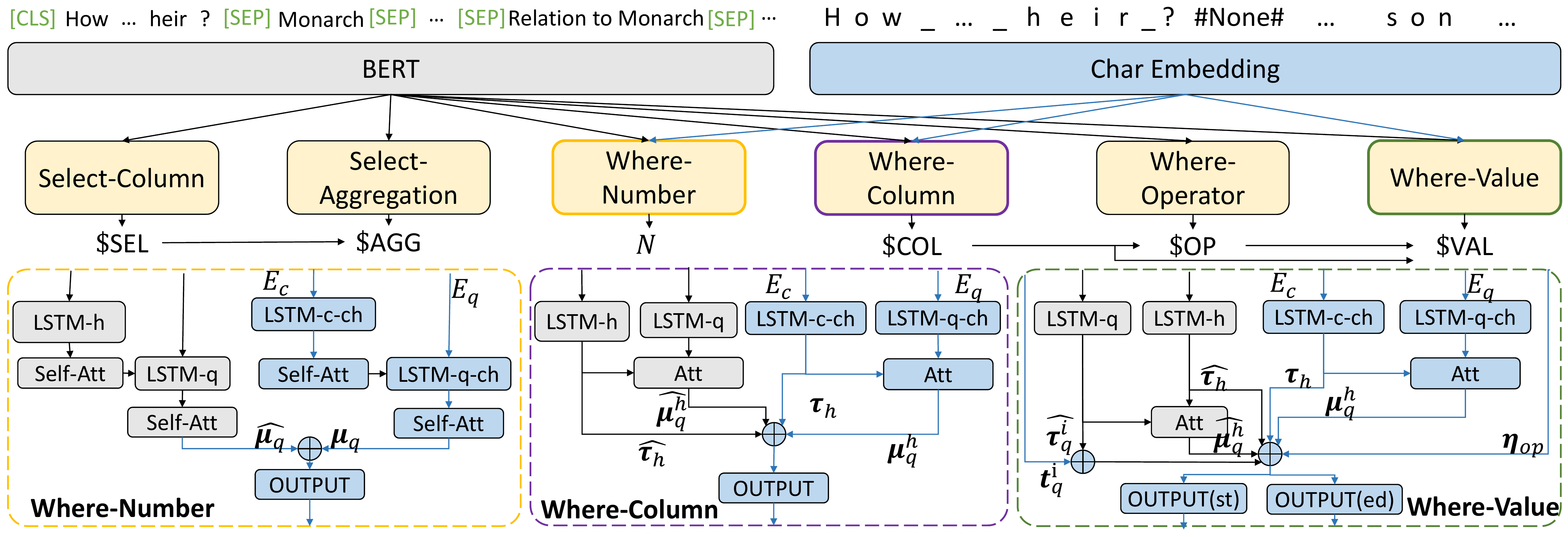}
		\caption{Architecture of the table content-enhanced model. WN, WC, and WV are detailed in the orange, purple, and green dotted box, respectively. Blue indicates the processes for table content and gray indicates the processes for headers.} \label{fig_model}
	\end{figure*}
	
	\section{Approach}
	The framework of our approach is shown in Figure \ref{fig_overview}, which consists of two parts. First, the \textit{table content enhanced model} (left) captures the semantic relevance of questions with headers and cells at the same time, and predict subtasks comprehensively. Further, \textit{zero-shot meta-learning} (right) is leveraged to train the table content enhanced model. In each training batch, the model parameters are updated in two stages to force the model to learn the generalization ability.
	
	\subsection{Table Content Enhanced Model}
	Figure \ref{fig_example} demonstrates that the core of leveraging table content is to find the table cells mentioned in the question. However, the total number of the cells can be very large, far exceeding that of the headers. Consequently, it is impractical to directly embed all of them. 
	
	To overcome this challenge, we adopt coarse-grained filtering before embedding. Specifically, for each header $h$, only the cell $c$ with the highest literal similarity to question $q$ will be retained. The literal similarity is computed by
	\begin{equation}
	{\rm \varphi}(c;q)=\max_{{\rm n}(q)} \frac{{\rm lcs}({\rm n}(q),c)}{2|{\rm n}(q)|}+\frac{{\rm lcs}({\rm n}(q),c)}{2|c|}
	\label{literal_similarity}
	\end{equation}
	where ${\rm n}(q)$ denotes the $n$-gram of $q$, $|x|$ denotes the length of string $x$, and ${\rm lcs}(x,y)$ denotes the length of the \textit{Longest Consecutive Common Subsequence} between the string $x$ and $y$. The intuitive meaning of ${\rm \varphi}(c;q) \in [0, 1]$ is that the larger proportion of the overlap in the two strings, the higher the similarity of them. In addition, if the retained cell whose score is smaller than the threshold $\sigma$, it will be replaced with a special token \#None\#, in order to avoid noise. After filtering, each header has a corresponding cell (or \#None\#).
	
	The overall architecture of our table content-enhanced model is shown in Figure \ref{fig_model}. It consists of an encoding module and six sub-modules corresponding to six sub-tasks. Intuitively, table content is mainly helpful to the sub-tasks for the WHERE clause, especially WN, WC, and WV. Therefore, we will detail these three sub-modules that utilize table content.
	
	\subsubsection{Encoding Module}
	The encoding module consists of BERT~\cite{DBLP:conf/naacl/DevlinCLT19} and an embedding layer. BERT is employed to encode the question and the headers. Following the format of BERT, the input is a sequence of tokens, which starts with a special token \texttt{[CLS]} followed by the word tokens of question $q$ and all headers $\{h^i\}$. A special token \texttt{[SEP]} is utilized to separate the $q$ and each $h^i$. For each input token, BERT outputs a hidden vector that holds its context information.
	In addition to BERT, the embedding layer is utilized to embed the cells. Differing from BERT, the embedding is for characters rather than words. It is because cells are typically some entity names and numerical values, etc. Char-embedding can reduce the number of Out-Of-Vocabulary (OOV) tokens. Specifically, For each cell $c$, its character embedding is denoted by $E_c \in \mathbf{R}^{m \times d_e}$, where $m$ is the character number of $c$ and $d_e$ is the dimension of each embedding. Since $q$ and $c$ should be embedded in the same vector space for calculating their semantic relevance, we also embed $q$ instead of directly using the BERT output of $q$. The char-embedding of $q$ is denoted by $E_q \in \mathbf{R}^{n \times d_e}$.
	
	\subsubsection{Where-Number Sub-Module}
	This sub-module contains two similar processes, which calculate the header-aware and content-aware context vector of the question, respectively. Here detail the process of the latter.
	First, for each $h$, its content vector $\bm{\tau}_h \in \mathbf{R}^{d}$ is obtained by a BiLSTM on $E_c$ followed by a max-pooling operation ($\mathbf{R}^{m \times d} \to \mathbf{R}^{d}$).
	Then, the hidden vectors of $q$ is obtained by the other BiLSTM, denoted by $\Gamma_q \in \mathbf{R}^{n \times d}$. In order to make this BiLSTM aware of $c$ when encoding $q$, its initial states $\bm{s} \in \mathbf{R}^{2d}$ is obtained by performing a self-attention on all $\bm{\tau}_h$.
	\begin{equation}
	\alpha_h = {\rm softmax}(\bm{\tau}_h W_{\alpha})
	\end{equation}
	\begin{equation}
	\bm{s} = W_s \sum_{h \in H}^n \alpha_h \bm{\tau}_h
	\end{equation}
	Here, $\alpha_h \in \mathbf{R}$ is the attention weight of each header $h$. $W_{\alpha} \in \mathbf{R}^{d \times d}$ and $W_{s} \in \mathbf{R}^{d \times 2d}$ are the trainable parameter matrices. Thereafter, the content-aware question context $\bm{\mu}_q \in \mathbf{R}^d$ is calculated by a self-attention on $\Gamma_q$, which is similar to $\bm{\mu}_h$. As described at the beginning, the header-aware question context $\widehat{\bm{\mu}}_q \in \mathbf{R}^d$ is calculated by the same procedures above on the output of BERT\footnote{In order to distinguish easily, all the vectors obtained from the output of BERT are marked with a hat, such as $\widehat{x}$.}. Finally, the result $\mathcal{N}$ is predicted by combining the header- and content-aware context vectors.
	\begin{equation}
	\mathcal{N} = \arg \max_{i} \tanh([\widehat{\bm{\mu}}_q;\bm{\mu}_q] W_{\mu} ) W_o
	\label{eq_wn}
	\end{equation}
	where $W_{\mu} \in \mathbf{R}^{2d \times d}$ and $W_o \in \mathbf{R}^{d \times 1}$ are the trainable parameter matrices.
	
	\subsubsection{Where-Column Sub-Module}
	In this sub-module,  the processes of calculating question hidden vectors $\Gamma_q$ and content vectors $\bm{\tau}_h$ are similar to those in WN.
	The only difference is the initial states of all the BiLSTMs are all random. Thereafter, in order to make the model focus on the parts of $q$ that are relevant to $c$, an attention mechanism is performed on $\Gamma_q$ to calculate content-aware context vector $\bm{\mu}^h_q \in \mathbf{R}^d$.
	\begin{equation}
	\bm{\mu}^h_q = \sum_{i=1}^n \alpha^i_q \bm{\gamma}^i_q
	\end{equation}
	\begin{equation}
	\alpha_q^i = {\rm softmax}(\bm{\gamma}^i_q W_{\alpha} {\bm{\tau}_h}^T)
	\end{equation}
	where $\bm{\gamma}^i_q \in \mathbf{R}^{d}$ is the $i$-th hidden vector in $\Gamma_q$ and $\alpha_q^i \in \mathbf{R}$ is its attention weight. $W_{\alpha} \in \mathbf{R}^{d \times d}$ is the parameter matrix. Finally, the result \$COL is predicted by
	\begin{equation}
	{\rm \$COL} = \arg \max_{h \in \mathcal{T}} \tanh([\widehat{\bm{\mu}}_q^h; \widehat{\bm{\tau}}_h; \bm{\mu}_q^h; \bm{\tau}_h]) W_o
	\end{equation}
	where $\widehat{\bm{\tau}}_h$ and $\widehat{\bm{\mu}}_q^h$ are the header vector and header-aware context vector, respectively. They are obtained by decoding the output of BERT with the same steps towards $\bm{\mu}_q^h$ and $\bm{\tau}_h$.
	
	\subsubsection{Where-Value Sub-Module}
	The architecture of this sub-module is almost consistent with that of WC. The difference is that prediction of \$VAL also requires the results of WC and WV, namely \$COL and \$OP. Let $h$ denote \$COL and $op$ denote \$OP, then the starting position $st$ and ending position $ed$ of \$VAL can be calculated by
	\begin{equation}
	st = \arg \max_{w^i \in q} \tanh([\widehat{\bm{\mu}}_q^h; \widehat{\bm{\tau}}_h; \bm{\mu}_q^h; \bm{\tau}_h; \bm{\eta}_{op}; \bm{\xi}_q^i]) W_{st}
	\end{equation}
	\begin{equation}
	ed = \arg \max_{w^i \in q} \tanh([\widehat{\bm{\mu}}_q^h; \widehat{\bm{\tau}}_h; \bm{\mu}_q^h;  \bm{\tau}_h; \bm{\eta}_{op}; \bm{\xi}_q^i]) W_{ed}
	\end{equation}
	Where $w^i$ is the $i$-th word token of $q$. $\widehat{\bm{\mu}}_q^h$, $\widehat{\bm{\tau}}_h$, $\bm{\mu}_q^h$, and $\bm{\tau}_h$ are calculated by the same procedures used in WC. $\bm{\eta}_{op}$ is the one-hot vector of \$OP and $\bm{\xi}_q^i$ is the semantic vector of $w^i$. Here, in order to leverage the content more directly, we propose a Value Linking(VL) strategy for calculating $\bm{\xi}_q^i$.
	\begin{equation}
	\bm{\xi}_q^i = [\widehat{\bm{\tau}}_q^i; \bm{t}_q^i]
	\label{eq_vl}
	\end{equation}
	where $\widehat{\bm{\tau}}_q^i \in \mathbf{R}^{d}$ is the hidden vector of $w^i$. It is calculated by a BiLSTM encoding the $q$ output of BERT. $\bm{t}_q^i \in \mathbf{R}^{d_{t}}$ denotes the type embedding of $w^i$. There are only two types, denoted by \texttt{Match} and \texttt{NotMatch}, that indicate whether $w^i$ matches some cells $c$, respectively. Initially, the type of each $w^i$ is labeled as \texttt{NotMatch}. When calculating the literal similarity by (\ref{literal_similarity}), if a cell $c$ is select, all the words of the corresponding n-gram $n(q)$ will be labeled as \texttt{Match}.
	
	The architectures of the remaining three modules SC, SA, and WO are almost consistent with WC, except that they do not need the process for table content (i.e., removing the blue process in Figure \ref{fig_model}). All their results are predicted by the classification that depends on the combined context $[\widehat{\bm{\tau}}_h; \widehat{\bm{\mu}}_q^h]$.
	
	\subsection{Zero-Shot Meta Learning Framework}
	Meta-learning is typically leveraged to deal with classification problems and has the ability to adapt quickly between different categories. In our proposed framework, the table that each sample (question) relies on is regarded as an abstract category, in order to create the conditions for applying meta-learning. Furthermore, the traditional meta-learning framework consists of two stages of meta-training and meta-test. However, it is already demonstrated in \cite{DBLP:conf/nips/VinyalsBLKW16,DBLP:conf/nips/SnellSZ17} where without fine-tuning on the meta-test, the meta-learning model shows similar even better performance. Motivated by this, our meta-learning algorithm only retains the meta-training stage. The entire process is formally described in Algorithm \ref{alg_meta_learning}.
	
	\begin{algorithm}
		\caption{Zero-Shot Meta-Learning Framework}
		\begin{algorithmic}[1]
			\label{alg_meta_learning}
			\REQUIRE A set of training samples $\mathcal{D}=\{(q^i, \mathcal{T}^i, y^i)\}$, where $q^i$ is the $i$-th input question, $t^i$ is the table which $q^i$ relies on, and $y^i$ is the gold SQL query of $q^i$. A model $\mathcal{M}(q,\theta)$, where $\theta$ is its parameters. Hyperparameters $\alpha$, $\beta$ and $\gamma$
			\WHILE{not done} 
			\FORALL{task}
			\STATE Sample a support set $\mathcal{S} = \{(q^j, \mathcal{T}^j, y^j)\} \subseteq \mathcal{D}$
			\STATE Evaluate $\nabla_{\theta}\mathcal{L}_{\mathcal{S}} = \nabla_{\theta}\Sigma_{j}\mathcal{L}(\mathcal{M}(q^j,\mathcal{T}^j,\theta), y^j)$
			\STATE Update parameters with gradient descent: \\$\phi = \theta - \alpha \nabla_{\theta}\mathcal{L}_{\mathcal{S}}$
			\STATE Sample a query set $\mathcal{Q} = \{(q^k, \mathcal{T}^k, y^k)\} \subseteq \mathcal{D}$, where $\{\mathcal{T}^j\} \cap \{\mathcal{T}^k\} = \emptyset$
			\STATE Evaluate $\nabla_{\theta}\mathcal{L}_{\mathcal{S} \gets \mathcal{Q}} =  \nabla_{\theta}\Sigma_{k}\mathcal{L}(\mathcal{M}(q^k,\mathcal{T}^k,\phi), y^k)$
			\STATE Update $\theta$ to minimum $\mathcal{L}$ using Adam optimizer with learning rate $\beta$, \\ where $\mathcal{L} = \gamma\mathcal{L}_{\mathcal{S}} + (1-\gamma)\mathcal{L}_{\mathcal{S} \gets \mathcal{Q}}$
			\ENDFOR
			\ENDWHILE
		\end{algorithmic}  
	\end{algorithm}
	A task consists of several training samples. It is the basic training unit of our framework and split into a support set and a query set. Here, to simulate the scenario of zero-shot tables, the table set of the support set is disjoint with that of the query set.  According to the split, the model experienced a two-stage gradient update during the training of each task. 
	In the first stage, temporary parameters $\phi$ are obtained by calculating the loss of the support set $\mathcal{S}$ and perform the gradient updating on original parameters  $\theta$. In the second stage, the loss of the query set $\mathcal{Q}$ is first calculated with $\phi$. Then, the losses of the support set and query set are jointed to calculate the gradient. Finally, original parameters $\theta$  are updated by the gradient.
	In addition, for sampling $\mathcal{S}$ and $\mathcal{Q}$, we follow the $N$-way $K$-shot setting, i.e, each set covers $N$ tables and there are $K$ samples for each table.
	
	Although meta-learning has also been utilized in \cite{DBLP:conf/naacl/HuangWSYH18} on text-to-SQL, there are two key differences between our proposed approach and their method: First, \cite{DBLP:conf/naacl/HuangWSYH18} focuses on sampling support sets according to types of the questions (e.g., COUNT, MIN), but we sample according to different tables, so as to capture the potential relationship between questions and tables. Second, we ensure that the tables in the support set do not intersect with those in the query set to simulate a zero-shot environment. Following this setting, the model needs to learn the generic knowledge between two different sets of tables and perform a joint optimization, thus it can be forced to learn the generalization ability.
	
	\section{Experiments}
	\subsection{Experimental Setup}
	Our models are trained and evaluated over the following two text-to-SQL benchmarks:
	
	\textbf{WikiSQL}~\cite{DBLP:journals/corr/abs-1709-00103} is an English open-domain text-to-SQL benchmark, containing more than 20K tables. Each question corresponds to a table, which is extracted from the Wikipedia page. The data set is divided into 56,355 training questions, 8,421 development questions, and 15,878 test questions\footnote{https://github.com/salesforce/WikiSQL}. In order to focus on evaluating the performance on zero-shot tables, we conduct experiments on the remaining 30\% of tables (zero-shot subset) released by \cite{DBLP:conf/aaai/ChangLT0HZ20}\footnote{https://github.com/JD-AI-Research-Silicon-Valley/auxiliary-task-for-text-to-sql}.
	
	\textbf{ESQL} is a Chinese domain-specific text-to-SQL dataset built by ourself. Its format imitates WikiSQL, containing 17 tables. These tables are related to the field of electric energy, including information such as electricity sales and prices\footnote{Due to commercial secrets, we first desensitize the original dataset and then release it and all the codes of MC-SQL on https://github.com/qjay612/meta\_learning\_NL2SQL, and all the results in this paper are obtained from the desensitized version.}, etc. Although the number of tables in ESQL is small, the number of headers in each table is several times that in a WikiSQL table, thus still covers a wealth of information. The dataset is divided into 10,000 training questions, 1,000 development questions, and 2,000 test questions. In order to simulate the challenge of zero-shot tables, the training set contains only 10 tables of all, while the development set and the test set contain all the tables. We respectively extract the questions from the development and test set that rely on the remaining 7 tables as the zero-shot subsets.
	
	Following previous approaches~\cite{DBLP:journals/corr/abs-1709-00103,DBLP:journals/corr/abs-1902-01069}, we adopt logical form (LF) accuracy and execution (EX) accuracy as the evaluation metrics. Here, LF evaluates the literal accuracy of the total SQL query and its clauses, and EX evaluates the accuracy of the results by executing the SQL query.
	
	\subsubsection{Implementation Details} 
	We perform all the experiments on NVIDIA Tesla V100 GPU. In the experiments, all the BERT models are of \textit{base} version. The following hyperparameters are tuned on development sets: (1) Fitlering threshold $\sigma$ is set to 0.9 for both datasets. (2) The layer number of all BiLSTMs is set to 2. (3) The hidden state size $d$ is set to 100. (4) The character embedding size $d_e$ is set to 128. (5) The type embedding size $d_t$ is set to 32. (6) The number of sampling tasks is set to 10,000 for WikiSQL, 2,500 for ESQL. (7) For WikiSQL, both $N$ and $K$ in the $N$-way $K$-shot setting are set to 4. For ESQL, they are set to 1 and 4, respectively. (8) $\gamma$ in Algorithm \ref{alg_meta_learning} is set to 0.3 for WikiSQL, 0.5 for ESQL. (9) For $\alpha$ in Algorithm \ref{alg_meta_learning}, BERT and sub-modules are trained with two kinds respectively. Specifically, $\alpha_{\rm BERT}$ is set to $1 \times 10^{-5}$ and $\alpha_{\rm sub}$ is set to $1 \times 10^{-3}$. Similarly, $\beta_{\rm BERT}$ is set to $1 \times 10^{-5}$ and $\beta_{\rm sub}$ is set to $1 \times 10^{-3}$.
	
	\subsection{Overall Results on WikiSQL}
	\begin{table}
		\begin{center}
			\scalebox{0.9}{
				\begin{tabular}{ccccc}
					\hline
					\rule{0pt}{12pt}
					Approach & Dev LF & Dev EX & Test LF & Test EX
					\\
					\hline
					\\[-6pt]	Seq2SQL & 49.5 & 60.8 & 48.3 & 59.4 \\
					Coarse2Fine & 72.5 & 79.0 & 71.7 & 78.5\\
					Auxiliary Mapping & 76.0 & 82.3 & 75.0 & 81.7 \\
					SQLova (-)& 80.3 & 85.8 & 79.4 & 85.2 \\
					SQLova (*) & 81.6 & 87.2 & 80.7 & 86.2 \\
					X-SQL (*) & 83.8 & 89.5 & 83.3 & 88.7 \\
					HydratNet (*) & 83.6 & 89.1 & \textbf{83.8} & 89.2 \\
					TaBERT-k1 (-) & 83.1 & 88.9 & 83.1 & 88.4 \\
					TaBERT-k3 (-) & 84.0 & 89.6 & 83.7 & 89.1 \\
					\hline
					\\[-6pt]
					MC-SQL (-) & \textbf{84.1} & \textbf{89.7} & 83.7 & \textbf{89.4} \\
					\hline
					\\[-6pt]
			\end{tabular}}
			{\caption{Overall results on WikiSQL. ``x(-)" denotes the model x with BERT-base. ``x(*)" denotes the model x with BERT-large or larger pre-trained model, such as MT-DNN~\cite{DBLP:conf/acl/LiuHCG19} in X-SQL. k1 and k3 indicate that the model considers 1 and 3 rows of related content for one question, respectively. }\label{tab_overall_wiksql}}
		\end{center}
	\end{table}

	\begin{table*}
		\begin{center}
			\scalebox{0.9}{
				\begin{tabular}{clccccccc}
					\hline
					\rule{0pt}{12pt}
					Dataset &Model &SC &SA &WN &WC &WO &WV &LF \\
					\hline
					\\[-6pt]
					\multirow{7}{*}{WikiSQL}
					&SQLova
					&96.7 / 96.3
					&90.1 / 90.3
					&98.4 / 98.2
					&94.1 / 93.6
					&97.1 / 96.8
					&94.8 / 94.3
					&80.2 / 79.7\\
					&TaBERT-k1
					&97.2 / \textbf{97.1}
					&90.5 / 90.6
					&98.9 / 98.8
					&96.1 / 96.1
					&\textbf{97.9} / \textbf{97.8}
					&\textbf{96.7} / 96.6
					&83.1 / 83.1\\
					&TaBERT-k3
					&\textbf{97.3} / \textbf{97.1}
					&\textbf{91.1} / \textbf{91.2}
					&98.8 / 98.7
					&96.6 / 96.4
					&97.5 / 97.5
					&96.6 / 96.2
					&83.9 / \textbf{83.7}\\
					&MC-SQL 
					&96.9 / 96.4
					&90.5 / 90.6
					&\textbf{99.1} / \textbf{98.8}
					&97.9 / \textbf{97.8}
					&97.5 / \textbf{97.8}
					&\textbf{96.7} / \textbf{96.9}
					&\textbf{84.1} / \textbf{83.7}\\
					&\quad w/o TC 
					&97.0 / 96.5
					&89.8 / 90.0
					&98.6 / 98.3
					&94.5 / 93.7
					&97.2 / 97.0
					&94.7 / 94.7
					&79.9 / 79.2\\
					&\quad w/o VL 
					&97.0 / 96.7
					&90.4 / \textbf{90.8}
					&99.0 / 98.7
					&\textbf{98.0} / 97.6
					&97.5 / 97.2
					&95.6 / 95.5
					&82.9 / 83.0\\
					&\quad w/o ML
					&96.5 / 96.2
					&90.4 / 90.4
					&98.9 / 98.7
					&97.8 / 97.4
					&97.5 / 97.4
					&96.5 / 96.1
					&83.2 / 82.9\\
					\\[-10pt]
					\hline
					\\[-6pt]
					\multirow{5}{*}{ESQL}
					&SQLova
					&96.2 / 95.9
					&98.9 / 99.0
					&98.5 / 98.4
					&84.6 / 84.1
					&96.5 / 95.8
					&89.9 / 89.6
					&72.0 / 71.5\\
					&MC-SQL 
					&\textbf{97.2} / \textbf{97.3}
					&99.1 / \textbf{99.2}
					&\textbf{98.9} / \textbf{98.9}
					&\textbf{93.6} / 93.3
					&\textbf{97.5} / 96.8
					&\textbf{92.9} / \textbf{92.6}
					&\textbf{82.8} / \textbf{82.7}\\
					&\quad w/o TC 
					&95.9 / 96.1
					&99.2 / 99.1
					&98.8 / 98.3
					&84.5 / 84.4
					&96.7 / 96.2
					&90.5 / 90.3
					&72.9 / 72.1\\
					&\quad w/o VL
					&96.5 / 96.7
					&\textbf{99.3} / 98.9
					&\textbf{98.9} / 98.8
					&93.5 / \textbf{93.5}
					&97.4 / \textbf{96.9}
					&92.0 / 91.8
					&82.1 / 81.9\\
					&\quad w/o ML
					&96.2 / 96.0
					&98.8 / 98.9
					&\textbf{98.9} / 98.8
					&92.4 / 92.7
					&\textbf{97.5} / 96.7
					&92.7 / 92.3
					&82.3 / 81.9\\
					\\[-10pt]
					\hline
					\\[-6pt]
					\multirow{7}{*}{\tabincell{c}{WikiSQL\\(zero-shot)}}
					&SQLova
					&95.8 / 95.2
					&89.7 / 89.3
					&97.6 / 97.4
					&91.1 / 90.4
					&95.9 / 95.7
					&90.1 / 90.5
					&74.7 / 72.8\\
					&TaBERT-k1
					&96.6 / \textbf{96.4}
					&91.0 / 91.0
					&98.6 / \textbf{98.4}
					&94.8 / 94.6
					&\textbf{97.7} / \textbf{97.5}
					&\textbf{95.3} / \textbf{94.6}
					&81.3 / 80.5\\
					&TaBERT-k3
					&\textbf{96.7} / \textbf{96.4}
					&\textbf{91.6} / \textbf{91.5}
					&98.2 / 98.2
					&95.1 / 95.0
					&96.8 / 97.0
					&94.9 / 94.2
					&82.0 / \textbf{81.2}\\
					&MC-SQL 
					&\textbf{96.4} / 95.5
					&91.1 / 91.0
					&\textbf{98.7} / 98.1
					&96.6 / \textbf{96.3}
					&97.1 / 96.7
					&94.8 / 94.2
					&\textbf{82.4} / 80.5 \\
					&\quad w/o TC
					&96.2 / \textbf{95.7}
					&91.0 / 90.5
					&97.6 / 97.7
					&91.5 / 90.7
					&96.2 / 96.1
					&90.5 / 90.8
					&75.8 / 73.6\\
					&\quad w/o VL
					&96.2 / 95.8
					&90.6 / 90.9
					&\textbf{98.7} / 98.0
					&\textbf{97.1} / \textbf{96.3}
					&97.1 / 96.3
					&91.7 / 92.1
					&79.0 / 79.1\\
					&\quad w/o ML 
					&95.7 / 95.0
					&90.4 / 90.2
					&98.5 / 98.2
					&96.0 / 95.8
					&96.8 / 96.7
					&94.0 / 93.5
					&81.2 / 79.4\\
					\\[-10pt]
					\hline
					\\[-6pt]
					\multirow{5}{*}{\tabincell{c}{ESQL\\(zero-shot)}}
					&SQLova
					&94.3 / 94.0
					&97.8 / 97.9
					&97.3 / 97.0
					&80.5 / 80.7
					&95.9 / 94.6
					&87.8 / 86.7
					&62.9 / 61.2\\
					&MC-SQL 
					&\textbf{94.6} / \textbf{94.2}
					&98.0 / 98.0
					&\textbf{97.5} / \textbf{97.3}
					&\textbf{93.7} / \textbf{92.0}
					&\textbf{96.2} / \textbf{94.8}
					&\textbf{91.9} / \textbf{90.5}
					&\textbf{76.7} / \textbf{74.8}\\
					&\quad w/o TC 
					&94.4 / 94.1
					&\textbf{98.1} / \textbf{98.2}
					&97.1 / 97.2
					&80.7 / 80.6
					&95.5 / 94.2
					&88.4 / 87.6
					&64.7 / 63.3\\
					&\quad w/o VL
					&93.8 / 94.0
					&98.0 / 98.1
					&97.4 / 97.2
					&92.6 / 91.1
					&95.1 / \textbf{94.8}
					&90.9 / 90.1
					&75.7 / 73.7\\
					&\quad w/o ML
					&93.5 / 93.0
					&97.7 / 97.9
					&97.4 / 96.9
					&93.2 / 91.8
					&96.0 / 94.3
					&91.2 / 90.2
					&75.2 / 72.9\\
					\\[-10pt]
					\hline
			\end{tabular}}
			{\caption{Results of sub-tasks on WikiSQL and ESQL. $x$/$y$ denotes the results of the dev/test sets.}\label{tab_submodule}}
		\end{center}
	\end{table*}
	
	We first compared our approach with several existing text-to-SQL approaches on public benchmark WikiSQL. Seq2SQL~\cite{DBLP:journals/corr/abs-1709-00103}, Coarse2Fine~\cite{DBLP:conf/acl/LapataD18}, and Auxiliary Mapping~\cite{DBLP:conf/aaai/ChangLT0HZ20} are all sequence-to-sequence (Seq2Seq) based models. SQLova~\cite{DBLP:journals/corr/abs-1902-01069} replaces the Seq2Seq framework with the multi-submodule framework.  X-SQL~\cite{DBLP:journals/corr/abs-1908-08113} and HydratNet~\cite{DBLP:journals/corr/abs-2008-04759} improve this framework by MT-DNN~\cite{DBLP:conf/acl/LiuHCG19} and a pair-wise ranking mechanism respectively, thus achieve better results. TaBERT~\cite{DBLP:conf/acl/YinNYR20} is a state-of-the-art language model used to encode tabular data. It is pre-trained on a massive number of question-tabular corpus and also uses content information. In this paper, we ignore all the results with execution guiding (EG) trick~\cite{DBLP:journals/corr/abs-1807-03100}. It is because EG works on the premise that the generated SQL query must not be empty, which is unreasonable in realistic scenarios.
	
	The overall experimental results on WikiSQL are reported in Table \ref{tab_overall_wiksql}. Except for TaBERT, where we use official API, all the other comparison results are directly taken from the original paper. On LF accuracy, our approach achieves state-of-the-art results on the development set and ranks second only to HydratNet (-0.1\%) on the test set. On EX accuracy, our approach achieves state-of-the-art results on both the sets. Notably, our results are achieved by only utilizing the \textit{base} version of BERT. After ignoring the baselines that use larger pre-trained models (``(*)" in Table \ref{tab_overall_wiksql}), our approach achieves significant improvements on both LF (4.3\%) and EX (4.2\%) accuracy when testing. In addition, compared with the table-specific pre-trained model, our model still has advantages without pre-training on table corpus. The performance of Seq2SQL, Coarse2Fine, and Auxiliary Mapping is limited by the decoding without SQL syntax constraints. SQLova, X-SQL, and HydratNet ignore the abundant information from table content, thus their performance is also limited. TaBERT makes use of content information and performs tabular-specific pre-training, thus achieving better results. There are two possible reasons why our approach outperforms TaBERT. On the one hand, TaBERT uses table information coarsely (SELECT-clause actually does not require content information), while we provide a more fine-grained usage (only WHERE-clause). On the other hand, the mandatory meta-learning process gives the model a stronger generalization ability.
	
	\subsection{Detailed Analysis}
	\subsubsection{Ablation Test} 
	To explore the contributions of various components of our MC-SQL model, we compared the following settings on both the datasets.
	\begin{itemize}
		\item \textbf{w/o table content(TC)} We removed all the processes in WN, WC, and WV that related to table content. For example, (\ref{eq_wn}) is converted to $p_{wn}(i | q, \mathcal{T}) = \tanh(\widehat{\bm{\mu}}_q W_{\mu} ) W_o$.
		\item \textbf{w/o value linking(VL)} We retained the processes related to TC but removed the value linking in WV, i.e., (\ref{eq_vl}) is converted to $\bm{\xi}_q^i = \widehat{\bm{\tau}}_q^i$ after removing.
		\item \textbf{w/o meta-learning(ML)} We replaced the meta-learning strategy with the traditional mini-batch strategy.
	\end{itemize}
	The detailed results on the full sets of WikiSQL and ESQL\footnote{The SQL query in ESQL also includes other keywords such as ORDER BY. We design modules similar to SELECT and WHERE above to solve them. The detailed results of these sub-tasks are available on the data set homepage.} are shown in the upper two blocks of Table \ref{tab_submodule}, respectively. MC-SQL equipped with all the components achieves the optimal results on LF accuracy and most sub-tasks, which significantly improves the baseline SQLova on both WikiSQL (4.0\%) and ESQL (10.2\%). Compared with TaBERT, our approach also has advantages on the overall performance on WikiSQL (0.2\% on Dev.) by the improvement on WC. This is probably due to the fine-grained use of content information for specific modules. By removing TC, the overall performance (LF) declined approximately 3.5\% and 10.6\% on both the datasets. It demonstrates the significance of content. Here, the performance drop by removing VL also proves the contribution of value linking. Removing ML brings a certain drop on both the datasets, however, the drop on ESQL (-1.8\%) is sharper than that on WikiSQL (-0.8\%). The reason can be that WikiSQL is an open-domain dataset, thus it is more difficult for generalization capability than domain-specific ESQL. By further observation, it can be found that the contribution of TC is mainly reflected in the four sub-tasks of WN, WC, and WV. The improvement on WO is mainly attributed to the improvement on WC, because the former depends on the result of the latter. In addition, meta-learning is helpful for all sub-tasks and has the most significant improvements on the sub-tasks that are not enhanced by table content, such as SC and SA. Interestingly, the performance sometimes becomes better on WC and WO after removing VL, which reveals that VL can be noisy for predicting \$COL. However, due to its significant improvement on WV, VL is still helpful for the overall performance.
	
	\subsubsection{Zero-shot Test} 
	We evaluated our model on the zero-shot subsets of both the datasets. The results are shown in the bottom two blocks of Table \ref{tab_submodule}. In terms of overall results, MC-SQL achieves greater improvements over SQLova on the zero-shot subsets of both WikiSQL (7.7\% vs 4.0\%) and ESQL (13.6\% vs 10.2\%). It proves that our approach is promising for enhance the model to handle zero-shot tables. Furthermore, the improvement on each subtask is also increased, especially WC and WV. The contribution of table content is greater on zero-shot tables, which is consistent with our intuition. The relatively more drastic performance drop caused by removing ML also proves that our meta-learning strategy is suitable for dealing with zero-shot tables. Notably, in addition to SC and SA, meta-learning is also contributing to the WHERE clause when handling zero-shot tables. It is interesting that the performance on ESQL is generally lower than that on WikiSQL, whereas the improvement brought by meta-learning on ESQL is greater than that of WikiSQL. We speculate that it is because fewer training tables result in poor performance, and meta-learning, which has the characteristic of suitable for a few samples, achieves greater improvements. Compared to TaBERT, our approach leads on the zero-shot development set (0.4\%) but lags behind on the zero-shot test set (-0.7\%). Specifically, TaBERT works better on the two sub-tasks of SA and WO. It benefits from its joint pre-training with tabular data and problems, thereby learning a stronger mapping capability between aggregation operators and questions.
	
	\subsubsection{Varied Sizes of Training Data} 
	\begin{figure}
		\includegraphics[width=0.45\textwidth]{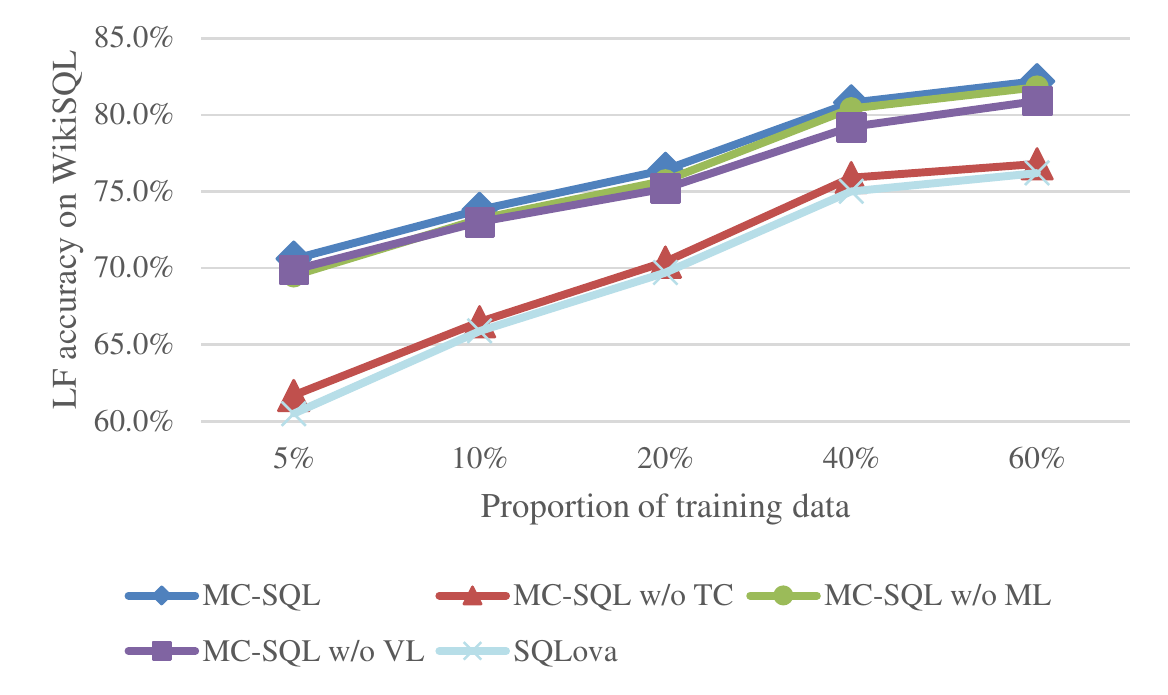}
		\caption{LF on WikiSQL with proportions of training data.} \label{fig_partial_wikisql}
	\end{figure}

	\begin{figure}
		\includegraphics[width=0.45\textwidth]{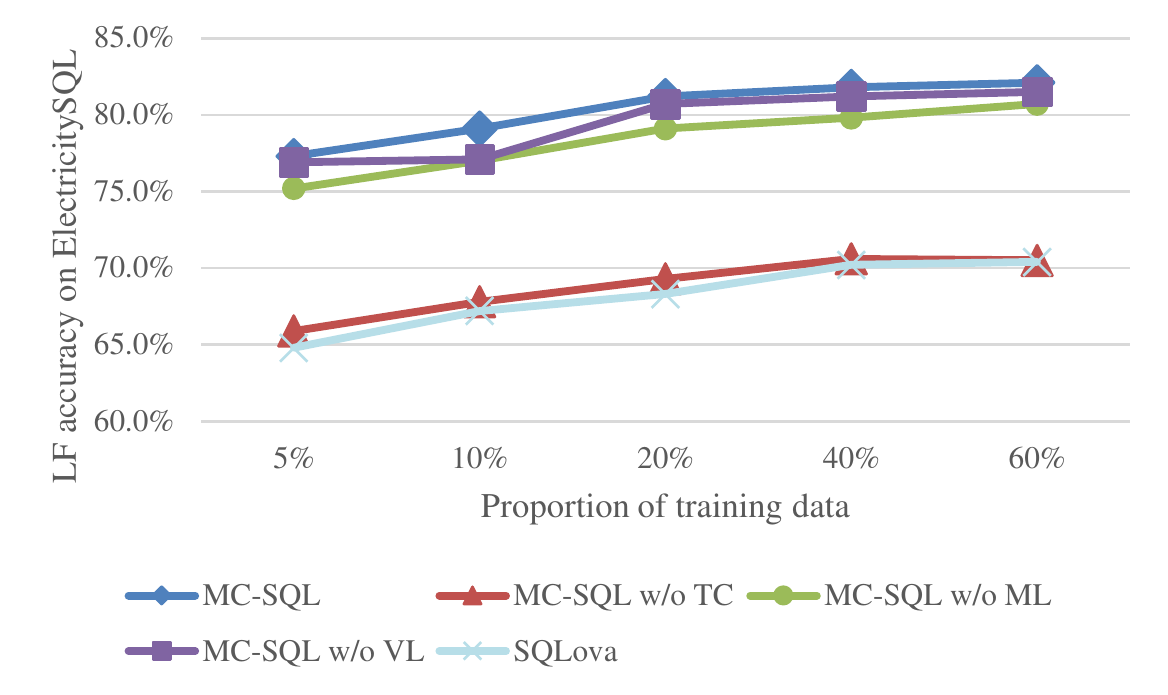}
		\caption{LF on ESQL with proportions of training data.} \label{fig_partial_esql}
	\end{figure}
	To simulate the scenario of zero-shot tables from another aspect, we tested the performance of the model using different proportions of training data. The results of WikiSQL and ESQL are shown in Figure \ref{fig_partial_wikisql} and Figure \ref{fig_partial_esql}, respectively. The MC-SQL equipped with all components always maintains optimal performance with different sizes of training data. When the training data is small, the improvement achieved by MC-SQL over SQLova is more significant, especially on WikiSQL. In addition, the results on both datasets demonstrate that the less training data, the more significant the improvement brought by meta-learning. Note that changes in training data have less impact on ESQL than that on WikiSQL. It is probably because of the few tables and the specific domain of ESQL.
	
	\section{Related Work}
	In recent research, mainstream text-to-SQL approaches mainly include two directions. 
	
	One direction is represented by Spider~\cite{DBLP:conf/emnlp/YuZYYWLMLYRZR18}, which is a benchmark to deal with the multi-table text-to-SQL task~\cite{DBLP:conf/naacl/YuLZZR18,DBLP:journals/corr/abs-1711-04436,DBLP:journals/corr/abs-1810-05237,DBLP:conf/acl/GuoZGXLLZ19,DBLP:conf/acl/WangSLPR20}. However, even if the evaluation does not require to recognize the value in WHERE clause, the state-of-the-art performance (65.6\% achieved by \cite{DBLP:conf/acl/WangSLPR20}) on this task is still far from realistic applications.
	
	The other direction is represented by WikiSQL~\cite{DBLP:journals/corr/abs-1709-00103}, which is a benchmark to deals with the single-table text-to-SQL task. This paper focuses on this task. Previous single-table text-to-SQL approaches~\cite{DBLP:journals/corr/abs-1709-00103,DBLP:journals/corr/abs-1711-04436,DBLP:conf/naacl/YuLZZR18,DBLP:conf/acl/LapataD18,DBLP:conf/aaai/ChangLT0HZ20} are mainly based on the Seq2Seq framework but ignore the characteristics of the SQL skeletons, thus their performance is limited. \cite{DBLP:journals/corr/abs-1902-01069} breaks the total task into several subtasks for the first time. They propose enables each sub-module to focus on the corresponding subtasks, thereby overcoming the bottleneck caused by a single model. In addition, the large-scale pre-trained language model~\cite{DBLP:conf/naacl/DevlinCLT19} also greatly improved model performance. Thereafter, almost all work on WikiSQL follows this framework of pre-trained models with multi-submodules. \cite{DBLP:journals/corr/abs-1908-08113} leverages the type information of table headers and replaces BERT with MT-DNN, which is a stronger pre-trained model trained from multi-task learning. \cite{DBLP:journals/corr/abs-2008-04759} proposes a pair-wise ranking mechanism for each question-header pair, and achieve better results on WikiSQL. The significant difference between our work and these approaches is that we take advantage of the content information. The closest work to ours is TABERT, which also encodes table content and utilizes a large number of question-table pairs for pre-training. However, their use of content lacks specificity, i.e., content-encoding is used for all subtasks. intuitively, the content information is helpful only for WHERE-clause predictions. Based on this intuition, our approach only uses content information on specific subtasks, thus it is more accurate. In addition, the use of meta-learning further promotes our model to obtain stronger generalization capabilities. More importantly, our approach can still play an important role in scenarios where lack of pre-trained models from massive tabular data (such as Chinese tables).
	
	\section{Conclusion}
	In this paper, we propose a new single-table text-to-SQL approach MC-SQL, which focuses on handling zero-shot tables. On the one hand, our approach takes advantage of table content to enhance the model. The potential header can be inferred by the semantic relevance of questions and content.  On the other hand, our approach learns the generalization capability from different tables by meta-learning. It utilizes a two-stage gradient update to force the model to learn generic knowledge. The experimental results show that our approach improves the baselines on multiple benchmarks. More importantly, the improvements further increase on zero-shot tables. In future work, we will try to classify different tables and combine meta-learning and reinforcement learning to further explore the generalization capabilities.
	
	\section{Acknowledgements}
	Research in this paper was partially supported by the National Key Research and Development Program of China under grants (2018YFC0830200, 2017YFB1002801), the Natural Science Foundation of China grants (U1736204), the Judicial Big Data Research Centre, School of Law at Southeast University.
	
	\bibliography{aaai21}
\end{document}